\documentclass[letterpaper, 10 pt, conference]{ieeeconf}  

\IEEEoverridecommandlockouts                              
\usepackage[utf8]{inputenc}
\usepackage{amsmath}
\usepackage{amssymb}
\usepackage{bm}
\usepackage{units}
\usepackage{comment}
\usepackage{url}

\usepackage{graphicx}
\usepackage{xcolor}
\usepackage{epstopdf}
\usepackage{pdfpages}
\usepackage{tikz}
\usepackage{pgfplots} 
\usepackage[footnotesize]{caption}
\usepackage{optidef}
\pgfplotsset{compat=1.12}
\usetikzlibrary{external}
\graphicspath{{./figures/}}
\definecolor{ao}{rgb}{0.0, 0.5, 0.0}

\newcommand{\pd}[2]{\frac{\partial #1}{\partial #2}}

\newcommand{\T}{^{\mathop{\mathrm{T}}}}

\newcommand{\vla}{\bm{\lambda}}

\newcommand{\vta}{\bm{\tau}}

\newcommand{\vb}{\mathbf b}
\newcommand{\vc}{\mathbf c}
\newcommand{\vd}{\mathbf d}

\newcommand{\vf}{\mathbf f}

\newcommand{\vp}{\mathbf p}
\newcommand{\vq}{\mathbf q}
\newcommand{\vr}{\mathbf r}

\newcommand{\vu}{\mathbf u}
\newcommand{\vv}{\mathbf v}

\newcommand{\vx}{\mathbf x}

\newcommand{\vA}{\mathbf A}

\newcommand{\vC}{\mathbf C}

\newcommand{\vI}{\mathbf I}
\newcommand{\vJ}{\mathbf J}
\newcommand{\vK}{\mathbf K}

\newcommand{\vP}{\mathbf P}

\newcommand{\cC}{\mathcal{C}}

\newcommand{\dx}{\dot{x}}
\newcommand{\dy}{\dot{y}}

\newcommand{\ramone}{RAM\textit{one}\,}

\newcommand{\rhex}{RHex\,}
\newcommand{\scarleth}{Scarl\textit{ETH}}

\newcommand{\anymal}{ANYmal}
\newcommand{\digit}{Digit}
\newcommand{\empower}{emPower}
\newcommand{\spacebok}{SpaceBok}
\newcommand{\norm}[1]{\left\lVert#1\right\rVert}

\newcommand{\inv}[1]{#1^{-1}}

\numberwithin{thm}{subsection}

\newcommand{\bbR}{\mathbb{R}}

\newcommand{\ISys}{}
\newcommand{\KSys}{}

\title{\LARGE \bf Bipedal walking with continuously compliant robotic legs}

\author{Robin Bendfeld and C. David Remy, \emph{Member, IEEE}
\thanks{This material is based on work supported by the Vector Stiftung. The authors thank the International Max Planck Research School for Intelligent Systems (IMPRS-IS) for supporting Robin Bendfeld.}
\thanks{The authors are with the Institute for Nonlinear Mechanics, Department of Mechanical Engineering, University of Stuttgart, Stuttgart, GE {\tt \small (bendfeld@inm.uni-stuttgart.de, david.remy@inm.uni-stuttgart.de)}}}

\begin{document}

\maketitle
\thispagestyle{empty}
\pagestyle{empty}
\begin{abstract}
In biomechanics and robotics, elasticity plays a crucial role in enhancing locomotion efficiency and stability.
Traditional approaches in legged robots often employ series elastic actuators (SEA) with discrete rigid components, which, while effective, add weight and complexity.
This paper presents an innovative alternative by integrating continuously compliant structures into the lower legs of a bipedal robot, fundamentally transforming the SEA concept.
Our approach replaces traditional rigid segments with lightweight, deformable materials, reducing overall mass and simplifying the actuation design.
This novel design introduces unique challenges in modeling, sensing, and control, due to the infinite dimensionality of continuously compliant elements.
We address these challenges through effective approximations and control strategies.
The paper details the design and modeling of the compliant leg structure, presents low-level force and kinematics controllers, and introduces a high-level posture controller with a gait scheduler. Experimental results demonstrate successful bipedal walking using this new design.
\end{abstract}

\section{Introduction}
In biomechanics, there is broad consensus that elasticity --achieved through the compliance of muscles, tendons, and ligaments~\cite{lichtwark2005invivo}-- provides significant benefits for legged locomotion by dampening impacts, tuning the dynamics of periodic movements, and storing and releasing energy throughout the gait cycle~\cite{alexander1990three}.
On a whole-body scale, the advantages of elasticity are clearly demonstrated by simple mechanical models, which capture the fundamental dynamics of legged locomotion.
The \textit{spring-loaded inverted pendulum} (SLIP) model~\cite{blickhan1989spring}, for example, uses a single massless spring to explain key aspects of human running.
When extended to two legs and equipped with a hip spring, this model can also describe the dynamics of many other bipedal gaits~\cite{gan2018all}.
Research has also shown that integrating springs can help robots embody desired natural dynamics~\cite{calzolari2023embodying}, and that they can facilitate the generation of energy-efficient gaits in bipeds~\cite{xi2014optimal, smit2017energetic} and quadrupeds~\cite{xi2016selecting, yesilevskiy2018energy}.

A widely adopted approach to incorporating elasticity in legged robotic systems is the use of \emph{series elastic actuation} (SEA)~\cite{pratt1995series}.
In SEA, an elastic element is placed between gearbox output and joint.
It protects the gearbox from impacts and decouples the motor’s reflected inertia from the joint.
Additionally, by measuring the deflection of the spring, joint torques can be estimated and regulated via admittance control.
When highly compliant elastic elements are used, they can periodically store and release energy during a stride, improving locomotion economy~\cite{hutter2012efficient}. 
Examples of electrically driven legged robotic systems that employ SEA include Anybotics' \anymal{}~\cite{hutter2016anymal}, Agility Robotics' \digit{}, the \empower{} bionic ankle, and \ramone{}~\cite{smit2017ramone}.

Despite the proven success of SEA, there is potential for significant improvements in its mechanical design.
Current SEA robots are constructed from rigid segments connected by joints, in which an additional elastic element is integrated alongside the motor.
This rigid structure adds mass to the robot's leg, which is further increased by the inclusion of the elastic component.
Moreover, co-locating motor, gearbox, and spring within a single rotational joint increases the overall design complexity.
\begin{figure}[t]
    \centering
    \includegraphics[width=\columnwidth]{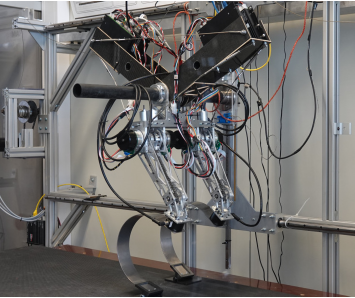}
    \caption{
    This paper demonstrates successful bipedal walking on continuously compliant legs.
    The hardware platform used for the experiments is based on the robot \ramone{}~\cite{smit2017ramone} in which the series compliance was removed from the actuators and the lower legs were replaced with a compliant semicircular structure made from spring steel.
    Bota Systems Rokubi F/T-sensors at the proximal end of the springs allow for measurement of the elastic forces.
    The robot is constrained to motion in the sagittal plane using a planarizer~\cite{green2016design}.
    }
    \label{fig:ramone-testbench}
    \vspace{-6mm}
\end{figure}

In our work, we propose an alternative approach that has the potential to fundamentally transform and enhance the SEA concept: we design the robot’s lower legs as deformable structures, allowing them to function as the series compliance element (Fig.\ref{fig:ramone-testbench}). 
This approach simplifies the actuation design significantly.
Additionally, by using lightweight, slender materials for these compliant segments, we can reduce the weight at the distal end of the leg, where it is particularly advantageous for improving efficiency and performance~\cite{browning2007effects}.
To draw the parallels to a traditional SEA, the rigid upper part of the leg with the hip and knee motors constitute the `actuator' which is positioned in series with the compliant lower leg which acts as the spring. 
Rather than controlling joint torques, however, we directly control the ground reaction forces at the foot.

While this design paradigm may offer advantages in terms of reduced weight, simplified complexity, and improved manufacturability, it also introduces substantial challenges in terms of modeling, sensing, and control of the continuously compliant elements.
Unlike traditional rigid robots, which rely on a finite number of discrete joints that can be easily equipped with sensors, a continuously compliant leg exhibits infinite dimensionality, necessitating appropriate approximations and simplifications for effective control.
The aim of this paper is to demonstrate how these challenges can be addressed and how a very simple controller can achieve stable bipedal locomotion using this innovative leg design.

Our work builds on several legged robots that integrate structural compliance into their designs.
\rhex{} used six fiberglass semicircular legs to navigate unstructured terrain~\cite{saranli2001rhex};
\spacebok{} was developed for lunar operations~\cite{arm2019spacebok};
an initial design of a two-legged robot designed for high-speed running using structural compliance was introduced in~\cite{niiyama2010athlete};
and one-legged hoppers with continuously compliant legs were introduced in~\cite{iida2012legged} and~\cite{zeglin1999bow}.
Although our design shares similarities with these previous robots, the motivation and application of continuous compliance in our system is distinct. 
\rhex{} utilized compliant legs primarily for obstacle negotiation and rough terrain traversal.
In \spacebok, the feet were made from carbon fiber for weight reduction (and later replaced with aluminum feet to enable locomotion on granular media~\cite{kolvenbach2021traversing}).
The one-legged hoppers demonstrated the use of resonance frequencies for energy-efficient locomotion.
In contrast, our approach uniquely employs the compliant legs as a series elastic element, actively regulating their deformation for force control.
This is an aspect not explored in the aforementioned robots.
This allows us to integrate compliance into the control architecture in a novel way, optimizing both stability and efficiency in legged locomotion.

We begin the remainder of this paper by detailing the new hardware (Section~\ref{sec:hardware-description}) and present suitable methods for modeling the compliance (Section~\ref{sec:mech-model}).
In Section~\ref{sec:control_policy}, we first introduce effective low-level controllers for force and kinematics regulation, before detailing a high-level posture controller supported by a gait scheduler.
Finally, the results of successful walking experiments, which confirm the overall viability of our proposed approach, are discussed in Section~\ref{sec:experiments}, and further demonstrated in the supplementary video.

\section{Hardware} \label{sec:hardware-description}
The experiments presented in this paper, were performed on a modified version of the bipedal robot \ramone{}~\cite{smit2017ramone}, which itself is based on the \scarleth{}~\cite{hutter2011scarleth} leg design.
As primary modifications, the elasticities in the hip and knee drives have been replaced by a rigid connection, such that the joints are driven directly by the Maxon EC60 flat motors and Harmonic Drives with a 50:1 gear ratio.
These motors are velocity regulated via EPOS3 motor controllers and equipped with encoders to measure joint angles and velocities.
In the following, we will denote the left (right) hip and knee angle with $\nu_{H,L}$ and $\nu_{K,L}$ ($\nu_{H,R}$ and $\nu_{K,R}$), respectively.
The robots's main body is constrained to move only in the sagittal plane by means of a sensorized motion planarizer~\cite{green2016design} that also measures the horizontal and vertical position $x_B$ and $y_B$, respectively.
The corresponding velocities are generated by forward-differentiating the positions and are low-pass filtered at \unit[20]{Hz}.
For the purpose of the experiments presented in this paper, we blocked the pitch of the main body.

The lower legs of the original \ramone{} design were replaced by curved elements made from \unit[500]{mm} long, \unit[2]{mm} thick and \unit[50]{mm} wide spring steel plates which were manufactured by Gutekunst Formfedern according to custom specifications (semi circle of radius \unit[120]{mm} with \unit[60]{mm} long straight attachments at both ends).
3D printed feet were mounted at the distal end of these elements.
They were made of an extended sole connected to the spring via a rotational passive ankle axis, similar to the feet in~\cite{daneshmand2021variable} and the robot Cassie~\cite{apgar2018fast}.
At the point of attachment at the knee, a Rokubi F/T-sensor from Bota Systems was placed between the knee adapter and the spring to record the elastic forces which were measured at \unit[1]{kHz} and filtered with a built-in filter at \unit[100]{Hz}.
No other sensors were integrated into the robotic leg.
In particular, contact is detected by thresholding the force values from the F/T-sensor instead of the inclusion of contact sensors in the feet.
The overall height of the robot is about \unit[685]{mm}, the overall mass \unit[10.9]{kg}.

\section{Modeling} \label{sec:mech-model}
In the following, we introduce a suitable model of this robot, focusing on the kinematics and the description of the compliant shank within a single leg.
\begin{figure}[t]
    \centering
    \includegraphics[width=\columnwidth]{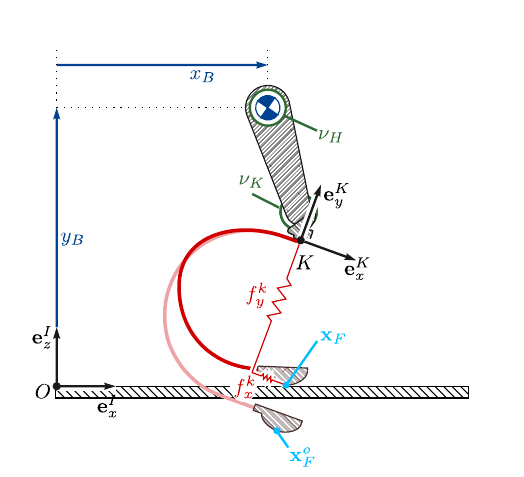}
    \caption{
    Model of a robotic leg with a deformable shank. 
    During the stance phase, the continuously compliant structure (shown in dark red) is deflected from its undeformed configuration (shown in light red).
    The elastic response is approximated by a linear elastic element which acts in response to the deflection of the foot from its nominal position (at $\vx_{F}^{o}$) to its actual position (at $\vq_F$).
    The underlying linear-elastic characteristics are constant in a spring-fixed coordinate system $K$, leading to restoring forces $f_x^k$ and $f_y^k$.
    (To improve readability, the sub-indices $i$ for left/right leg have been omitted in this figure)
    }
    \vspace{-6mm}
    \label{fig:fig-compliant-leg-model}
\end{figure}
\subsection{Generalized coordinates and torques}
Using the quantities introduced in Sec.~\ref{sec:hardware-description}, we define the vector $\vx_B\T = \begin{bmatrix}\ISys{}x_B & \ISys{}y_B\end{bmatrix}$ to denote the motion of the main body and the vector $\vq_{\nu}\T = \begin{bmatrix}\vq_{\nu,L}\T & \vq_{\nu,R}\T\end{bmatrix}$ with $\vq_{\nu,i}\T = \begin{bmatrix}\nu_{H,i} & \nu_{K,i}\end{bmatrix}$ (where $i \in \{ L, R\}$ denotes the left and right leg, respectively) to describe the motion of the actuated degrees of freedom.
To describe the deflection of the elastic shank and to complete the vector of generalized coordinates $\vq\T = \begin{bmatrix} \vx_B\T & \vq_{\nu}\T & \vx_{F}\T \end{bmatrix}$, we use the position of the contact point at the feet $\vx_{F}\T = \begin{bmatrix}\vx_{F,L}\T & \vx_{F,R}\T\end{bmatrix}$ with $\vx_{F,i}\T = \begin{bmatrix}\ISys{}x_{F,i} & \ISys{}y_{F,i}\end{bmatrix}$.
This choice of coordinates will allow us later to set $\dot{\vx}_{F,i}=\mathbf{0}$ during stance, assuming that the corresponding foot is not sliding.
In addition, the corresponding generalized forces are the ground reaction forces $\ISys{}\vla\T = \begin{bmatrix}\ISys{}\vla_L\T & \ISys{}\vla_R\T\end{bmatrix}$.
The full set of generalized forces $\vta$ is given as $\vta\T = \begin{bmatrix} \ISys{}\vta_B\T & \vta_{\nu}\T & \ISys{}\vla\T \end{bmatrix}$, where $\vta_{\nu}\T =\begin{bmatrix}\vta_{\nu,L}\T & \vta_{\nu,R}\T\end{bmatrix}$ are the motor torques and $\ISys{}\vta_B$ the forces that the leg applies to the main body.

The nominal position of a foot $\ISys{}\vx_{F,i}^{o}$ (i.e., the position where the foot would be located for an undeformed elastic element) is expressed via the kinematic function
\begin{equation} \label{eq:forward-kinematics-pos}
    \ISys{}\vx_{F,i}^{o}(\vq) = \vq_B  + \vr_{i}^{kin}(\vq_{\nu,i}) \,,
\end{equation}
with the time derivative
\begin{equation} \label{eq:forward-kinematics-vel}
    \ISys{}\vv_{F,i}^{o}(\vq,\dot{\vq}) = \dot{\vq}_B  + \vJ_{\nu,i}(\vq_{\nu,i})\dot{\vq}_{\nu,i} \,,
\end{equation}
using the actuator Jacobian $\vJ_{\nu,i} = \pd{\vr_{i}^{kin}(\vq_{\nu,i})}{\vq_{\nu,i}}$.
Based on this nominal position, we can compute the deflection of the elastic element as:
\begin{equation} \label{eq:delta}
    \ISys{}\Delta\vx_{F,i}(\vq) = \ISys{}\vx_{F,i} - \ISys{}\vx_{F,i}^{o} = \vx_{F,i} - \vq_B - \vf_{i}^{kin}(\vq_{\nu,i})   \,.
\end{equation}
The overall Jacobian associated with this deflection is $\vJ_{i}(\vq) = \pd{\ISys{}\Delta\vx_{F,i}}{\vq}$, which is given for our choice of generalized coordinates as $\vJ_{L} = \begin{bmatrix} -\vI_{2 \times 2} & -\vJ_{\nu,L} & \mathbf{0}_{2 \times 2} & \vI_{2 \times 2} & \mathbf{0}_{2 \times 2} \end{bmatrix}$ for the left leg and as $\vJ_{R} = \begin{bmatrix} -\vI_{2 \times 2}& \mathbf{0}_{2 \times 2} & -\vJ_{\nu,R} & \mathbf{0}_{2 \times 2} & \vI_{2 \times 2} \end{bmatrix}$ for the right leg.
All quantities introduced so far are defined in the inertial $I$-frame of reference.

\subsection{Model of the compliance during the stance phase}
In contrast to our early work in this area, which used a Hencky-type beam formulation with multiple independent degrees of freedom for modeling~\cite{bendfeld2023modeling} and control~\cite{bendfeld2023contact}, we showed in more recent work~\cite{bendfeld2024squatting} that a simplified mechanical model which approximates the elasticity by a linear mass-free spring can vastly reduce the complexity of the controller development and has sufficient prediction capabilities in the region of deformation a single leg can achieve when performing squatting and locomotion tasks.

To facilitate this representation, we introduce a moving `spring'-frame of reference which is denoted by $K_i$ (Fig.~\ref{fig:fig-compliant-leg-model}).
This frame is fixed to the adapter at the knee of the two legs.
The transformation to the inertial $I$-frame is given by the transformation matrix $\vA_{IK_{i}}(\vq_{\nu,i})$.
In the $K_{i}$-coordinate systems the linear elasticity model has constant coefficients and is given by the symmetric positive definite stiffness matrix:
\begin{equation}\label{eq:fit-stiffness-matrix}
    \vK_i = \begin{bmatrix}
        k_{xx,i} & k_{xy,i} \\
        k_{xy,i} & k_{yy,i}
    \end{bmatrix} \,.
\end{equation}
With this elasticity model, the resulting elastic forces $\ISys{}\vf_i^k(\vq)$ are given in $I$-coordinates as:
\begin{equation}\label{eq:elastic-forces}
    \ISys{}\vf_i^k(\vq) = \vA_{IK_i}(\vq_{\nu,i}) \, \KSys{}\vK_i \, \vA_{K_iI}(\vq_{\nu,i}) \ISys{}\Delta\vx_{F,i}(\vq) \,.
\end{equation}

With the transposed of the Jacobian, the forces from both legs are projected back into the generalized coordinates~\cite{glocker2013set}:
\begin{equation}\label{eq:restoring-force-law}
    \vta^k(\vq) = \vJ_L\T(\vq) \ISys{}\vf_L^k (\vq) + \vJ_R\T(\vq) \ISys{}\vf_R^k (\vq)\,.
\end{equation}
They can be separated into their specific contributions to the different degrees of freedom:
\begin{equation}\label{eq:static-equilibrium-kframe}
            \vta^k(\vq) =
            \begin{bmatrix} \vta_{B}^k \\ \vta_{\nu,L}^k \\ \vta_{\nu,R}^k\\ \vla_L^k \\ \vla_R^k \end{bmatrix} = 
            \begin{bmatrix}
                -\ISys{}\vf_L^k (\vq) - \ISys{}\vf_R^k (\vq) \\
                -\vJ_{\nu,L}\T(\vq) \ISys{}\vf_L^k (\vq) \\
                -\vJ_{\nu,R}\T(\vq) \ISys{}\vf_R^k (\vq) \\
                \ISys{}\vf_L^k (\vq) \\
                \ISys{}\vf_R^k (\vq) \\
            \end{bmatrix}\,.   
\end{equation}

\section{Control}\label{sec:control_policy}
In the following section, we introduce our walking controller that relies on low-level controllers for the regulation of contact forces and foot point positions, algortihms for the control of the main body and swing leg, and an overarching gait scheduler. 
The inputs to the plant, i.e., to the EPOS 3 motor controllers, are desired motor velocities measured values are joint kinematics and the force readings from the F/T-sensor.

\subsection{Low-level position control}\label{ssec:low-level-position-control}
During the swing phase, a foot contact point must follow a swing foot trajectory, that is obtained from the high-level controller as a desired foot position $\ISys{}\vx_{F,i}^d(t)$ and velocity $\ISys{}\vv_{F,i}^d(t)$.
To this end, we employ a kinematic controller that is based on the simplifying assumption that the spring is not undergoing any deflection during swing.
In other words, we simply equate $\ISys{}\vx_{F,i}^{o}(\vq^d) = \ISys{}\vx_{F,i}^d$ and $\ISys{}\vv_{F,i}^{o}(\vq^d,\dot{\vq}^d) = \ISys{}\vv_{F,i}^d$ and perform inverse kinematics on eqs.~\eqref{eq:forward-kinematics-pos}~\&~\eqref{eq:forward-kinematics-vel} to solve for the desired joint angles $\vq_{\nu,i}^d$ and velocities $\dot{\vq}_{\nu,i}^d$ under given desired main body trajectories $\vq_{B}^d$ and $\dot{\vq}_{B}^d$.
From these, the inputs $\bm{u}_{i}$ (i.e., the velocities commanded to the EPOS$3$ motor controllers of this leg) are computed via the P-controller:
\begin{equation}\label{eq:actuated-joint-position-control}
	\bm{u}_{i} = \dot{\vq}_{\nu,i}^d + \vP_{\nu}(\vq_{\nu,i}^d-\vq_{\nu,i})\,,
\end{equation}
with the positive definite gain matrix $\vP_{\nu}$.

\subsection{Low-level force control}\label{ssec:low-level-force-control}
When a foot is on the ground, our control goal is to regulate the contact forces $\vla_{i}$.
These are provided from the high-level controller as desired values $\vla_{i}^{d}$ with known time derivatives $\dot{\vla}_{i}^{d}$.
Overall, we follow our controller design described in~\cite{bendfeld2023contact} and~\cite{bendfeld2024squatting}.

Assuming that the elastic segment and foot have negligible mass, we can equate the actual contact forces $\vla_{i}$ with the projection of the elastic forces into $\vq_{F,i}$; that is, with $\vla_{i}^k$ from eq.~\eqref{eq:static-equilibrium-kframe}.
These elastic forces are given by
\begin{equation}\label{eq:spring-influence-on-foot-coordinates2}
    \vla_{i}^k = \vA_{IK_{i}} \, \vK_{i} \, \vA_{K_{i}I} \big(\vq_{F,i} - \vq_B - \vr_{i}^{kin}(\vq_{\nu,i}) \big)\,.
\end{equation}
To arrive at a control law with suitable inputs, we must consider the time derivative of eq.~\eqref{eq:spring-influence-on-foot-coordinates2}:
\begin{equation}
    \dot{\vla}_{i}^k = \pd{\vla_{i}^k}{\vq}\dot{\vq} = \pd{\vla_{i}^k}{\vq_B}\dot{\vq}_B + \pd{\vla_{i}^k}{\vq_{\nu,i}}\dot{\vq}_{\nu,i} + \pd{\vla_{i}^k}{\vq_{F,i}}\dot{\vq}_{F,i}\,.
\end{equation}
Assuming non-slipping of the foot ($\dot{\vq}_{F,i} = \bm{0}$) and substituting $\bm{g}_{i}(\vq) = \pd{\vla_{i}^k}{\vq_{\nu,i}}$ and $\bm{w}_B(\vq) = \pd{\vla_{i}^k}{\vq_B}$, this simplifies to:
\begin{equation}\label{eq:rate-of-change-force-final}
    \dot{\vla}_{i}^{k} = \bm{g}_{i}(\vq)\dot{\vq}_{\nu,i} + \bm{w}_B(\vq)\dot{\vq}_{B}\,.
\end{equation}
To track a desired contact force trajectory $\vla_{i}^{d}(t)$ (i.e., \mbox{$\lim_{t\rightarrow \infty} \vla_{i}^k(t) - \vla_{i}^d(t) = \bm{0}$}, we design stable error dynamics:
\begin{equation}\label{eq:stable-error-dynamics}
    \big(\dot{\vla}_{i}^{k}-\dot{\vla}_{i}^d\big) + \vP_\lambda\big(\vla_{i}^k - \vla_{i}^d\big) = \bm{0}\,,
\end{equation}
with the positive definite gain matrix $\vP_{\nu}$.

By equating eqs.~\eqref{eq:rate-of-change-force-final}~\&~\eqref{eq:stable-error-dynamics}, substituting $\dot{\vq}_{B}$ with the desired value $\dot{\vq}_{B}^{d}$, solving for $\dot{\vq}_{\nu,i}$ and using this velocity as the input $\vu_{i}$ to the motor controllers of this leg, we arrive at the final force controller: 
\begin{equation} \label{eq:direct-force-controller}
    \bm{u}_{i} = \inv{\bm{g}_{i}}(\vq) \Big(\dot{\vla}_{i}^d + \vP_\lambda\big(\vla_{i}^d-\vla_{i}^k\big) - \bm{w}_B(\vq)\dot{\vq}_{B}^{d}\Big) \,. 
\end{equation}
This controller includes a feed-forward term based on $\dot{\vla}_{i}^d$, a feedback term using data from the F/T-sensors to obtain measurements of $\vla_{i}^k$, and another feed-forward term to account for the main body motion $\dot{\vq}_{B}^{d}$.
Note that in order to apply the controller from eq.~\eqref{eq:direct-force-controller}, we have to ensure that $\bm{g}_{i}(\vq)$ is invertible~\cite{bendfeld2023contact}.
In particular, we have to stay clear of kinematic singularities.

\subsection{Main body control}\label{ssec:ps_control}
The goal of the main body controller is to maintain a desired height $\ISys{}^{}y_B^d(t)$ and desired horizontal position $\ISys{}^{}x_B^d(t)$ of the robot's main body within the inertial frame of reference.
Similar to~\cite{pratt2001virtual}, we use the following simple trajectories for the main body:
\begin{equation}\label{eq:com_traj_simple}
        \ISys{}^{}x_B^d(t) = \ISys{}^{}v_B^d t \,, \quad 
        \ISys{}^{}y_B^d(t) \equiv \text{const.}  \,,
\end{equation}
where the constant $\ISys{}^{}v_{B}^d$ denotes the desired horizontal velocity.

To stabilize the main body we employ a \textit{virtual model controller}~(VMC)~\cite{pratt2001virtual}. This controller is easily parametrizable, robust against disturbances and does not need a precise dynamical model of the robot. 
In this controller, PD-laws (thought of as virtual springs and dampers) compute desired forces ${\ISys{}^{}\vta_B^d}\T = \begin{bmatrix}\ISys{}^{}\tau_x^\mathrm{d} & \ISys{}^{}\tau_y^\mathrm{d}\end{bmatrix}$ that act on the main body. They are given in horizontal and vertical direction as
\begin{align}
	\tau_x^\mathrm{d} &= p_x(x_B^d-x_B) + d_x(\dx_B^d-\dx_B) \\
	\tau_y^\mathrm{d} &= p_y(y_B^d-y_B) + d_y(\dy_B^d-\dy_B) + m_B g
\end{align}
with the non-negative proportional gains (virtual spring constants) $p_x$, $p_y$ and the derivative gains (virtual damping elements) $d_x$ and $d_y$.
The term $ m_B g$ is for gravity compensation, where $m_B$ is the mass of the main body and $g$ the gravitational constant.
Please note that for notational simplicity we omit the explicit notion of time dependence in the desired and measured states.

These desired forces $\vta_B^d$ are created through the ground reaction forces of both legs, which act additively:
\begin{equation}
	\vta_B^d = \vla_L^{d} + \vla_R^{d}\,.
\end{equation}
To distribute the desired forces between the two legs, we solve a constraint optimization problem, which minimizes the differences between desired and actual forces while minimizing the horizontal components of the ground reaction forces (with a penalty weight of $w_{x}$).
It further constraints the vertical forces to be within an admissible range (between $0$ and $\lambda_{y,i}^{\mathrm{max}}$) and the horizontal forces to lie within the friction cone with a nominal coefficient of friction $\mu$:
\begin{mini}|s|
{\vla^d\in\bbR^4}{\frac{1}{2}\norm{\vA\vla^d-\vb}_2^2}{}{}
\addConstraint{|{\mu\lambda_{x,L}^d|} \leq \lambda_{y,L}^d}
\addConstraint{|{\mu\lambda_{x,R}^d|} \leq \lambda_{y,R}^d}
\addConstraint{0\leq\lambda_{y,L}^d\leq\lambda_{y,L}^{\mathrm{max}}}
\addConstraint{0\leq\lambda_{y,R}^d\leq\lambda_{y,R}^{\mathrm{max}}}
\label{opt:force-distribution-i}
\end{mini}
with
\begin{equation*}
    \vA = \begin{bmatrix}
        1 & 0 & 1 & 0 \\
        0 & 1 & 0 & 1 \\
        w_{x} & 0 & 0 & 0 \\
        0 & 0 & w_{x} & 0
    \end{bmatrix} \,,\, \vb = \begin{bmatrix}
        \tau_x^d \\ \tau_y^d \\ 
        0 \\ 0 
    \end{bmatrix}\,.
\end{equation*}
We rewrite~\eqref{opt:force-distribution-i} into the quadratic program:
\begin{mini}|s|
{\vla^d\in\bbR^4}{\frac{1}{2}(\vla^d)\T\bar{\vA}\vla^d + \bar{\vb}\T\vla^d}{}{}
\addConstraint{\bar{\vC}\vla^d \leq \bar{\vd}}
\label{opt:force-distribution-ii}
\end{mini}
with the quantities
\begin{equation*}
    \bar{\vA} = \vA\T\vA \,, \bar{\vb} = -\vA\T\vb 
\end{equation*}
and
\begin{equation*}
    \bar{\vC} = \begin{bmatrix}
        \mu & -1 & 0 & 0 \\
        -\mu & -1 & 0 & 0 \\
        0 & 0 & \mu & -1 \\
        0 & 0 & -\mu & -1 \\
        0 & 1 & 0 & 0 \\
        0 & 0 & 0 & 1
    \end{bmatrix}\,, \vd = \begin{bmatrix}
        0 \\ 0 \\ 0 \\ 0 \\ \lambda_{y,L}^{\mathrm{max}} \\ \lambda_{y,R}^{\mathrm{max}}
    \end{bmatrix}\,.
\end{equation*}
The vertical force limits $\lambda_{y,i}^{\mathrm{max}}$ are also used to deactivate a leg during the swing phase.
For the single support phases, we choose \mbox{$\lambda_{y,L}^{\mathrm{max}} = $ \unit[0]{N}} for right leg single support and \mbox{$\lambda_{y,R}^{\mathrm{max}} = $ \unit[0]{N}} for left leg single support.

\subsection{Swing leg control}\label{ssec:swing_control}
During swing, the desired foot trajectory is defined as a two-segment piece-wise cubic polynomial $\vp^{d}$ which is expressed relative to the robots main body.
This kinematic function is added to the desired main body positions to yield foot point trajectories in the inertial frame of reference:
\begin{equation}\label{eq:swing_foot_traj}
    \vx_{F,i}^d(t) = 
	\begin{bmatrix}
		x_{F,i}^d \\ y_{F,i}^d
	\end{bmatrix} = \vq_B + \vp_i^d\left(s_{\mathrm{swi},i}(t),\vq\right)\,.
\end{equation}
The timing variable $s_{\mathrm{swi},i}(t)$ is a normalized swing time, provided by the gait scheduler (see below).
The trajectory itself is designed to fulfill the following criteria:
\begin{itemize}
    \item Trajectory starts at current foot position (hence the dependencey on $\vq$).	
    \item Near vertical lift-off with velocity $\dy_0^d$ to prevent scuffing of the foot.
    \item Apex transit at a given height given of $h_{\mathrm{swi}}$.
	\item Touchdown at the end of swing with velocity $\dy_f^d$ after the required horizontal travel.
	\item Trajectory is kinematically feasible; i.e. $\begin{bmatrix}
		x_{F,i}^d & y_{F,i}^d
	\end{bmatrix} \in \cC_i$, where $\cC_i$ is the configuration space of leg $i$.
\end{itemize}

\subsection{Gait scheduling} \label{sssec:gait_scheduling}
The gait schedule is defined by the stride time $T_\mathrm{str}$, the duty factor $\beta \in \left(0,1\right)$ (which determines the fraction of time each leg is on the ground), and the relative phase value of each leg $\theta_i \in \left[0,1\right)$.
For walking, we would set  $\theta_L = 0$ and $\theta_R=0.5$, with a duty factor  $\beta > 0.5$.

Given the time of walking $t\geq 0$, we first compute a normalized stride time $s_\mathrm{str}$ according to:
\begin{equation}\label{eq:s_stride}
	s_\mathrm{str}(t) = \frac{t \pmod{T_\mathrm{str}}}{T_\mathrm{str}} \in \left[0,1\right)\,.
\end{equation}
Based on this quantity, each leg is assigned an individual leg time $s_i \in \left[0,1\right)$, which is defined as
\begin{equation}\label{eq:s_foot}
	s_i(s_\mathrm{str}) = \begin{cases}
		s_\mathrm{str} - \theta_i & s_\mathrm{str} - \theta_i \geq 0 \\
		s_\mathrm{str} - \theta_i + 1 & s_\mathrm{str} - \theta_i < 0 \\
	\end{cases}
\end{equation}
Using this leg time, $s_i = 0$ corresponds to mid-stance, swing is scheduled to start at $\frac{\beta}{2}$, and swing should end at $1-\frac{\beta}{2}$.
Finally, we define a normalized swing time:
\begin{equation}\label{eq:swing_time}
	s_{\mathrm{swi},i} = \begin{cases}
		0 & s_i < \frac{\beta}{2}\,, \\
		1 & s_i > 1-\frac{\beta}{2}\,, \\
		\frac{s_i-\frac{\beta}{2}}{1-\beta} & \text{otherwise,}
	\end{cases}
\end{equation}
which runs from $0$ to $1$ during the swing phase and is used in the swing leg controller described in Sec.~\ref{ssec:swing_control}.

Two independent state machines determine the phase $c_i \in \{0,1\}$ of each leg.
Here, the swing phase corresponds to $c_i = 0$ and the stance phase to $c_i = 1$.
The transition from stance to swing is purely time based, and triggered at $s_i = \frac{\beta}{2}$.
The transition back to into the stance phase is based on actual measured values of the F/T-sensors which need to exceed a value of $\lambda_{y}^{min}>$ \unit[5]{N} to prevent the use of torque control, while the leg is in the air.
However, the swing leg trajectories are designed that the swing leg is brought back into contact at around $s_i = 1- \frac{\beta}{2}$.
The gait is initialized with $\vc = \bm{1}$.  That is, all legs start in stance phase.

\subsection{Overall controller} \label{sssec:allControl}
Within each controller execution, we first set $\lambda_{y,i}^{\mathrm{max}}$ to $0$ if $c_i = 0$ indicates that the leg is in swing.
We then compute $\vla^{d}$ according to eq.~\eqref{opt:force-distribution-ii}.
Finally, we use eq.~\eqref{eq:direct-force-controller} to compute the motor velocity commands of the legs that are in stance ($c_i = 1$) and eq.~\eqref{eq:actuated-joint-position-control} to compute the motor velocity commands of the legs that are in swing ($c_i = 0$).

\section{Experiments} \label{sec:experiments}
To produce walking, we commanded a desired constant velocity of $v_{B}^{d} = $ \unitfrac[0.15]{m}{s} and a desired constant height of $y_B^{d} = $ \unit[0.62]{m}.
The gait scheduling parameters were chosen as $T_{\mathrm{str}} = $ \unit[1.2]{s}, $\beta = 0.6$,  $\theta_L = 0$, and $\theta_R=0.5$. 
The controller gains and remaining parameters are given in Table~\ref{tab:control-parameters}. 
\begin{figure}[t]
    \centering
    \includegraphics[width=\columnwidth]{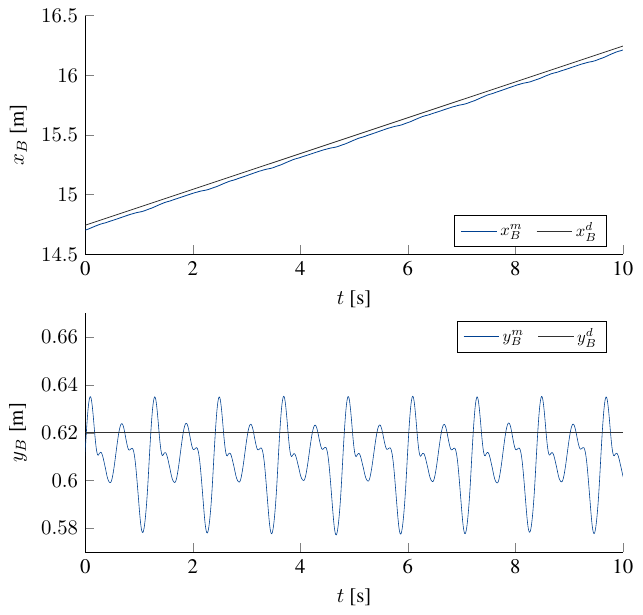}
    \caption{Shown are 10 seconds of the vertical and horizontal positions of the main body of the bipedal robot during a steady state walking experiment is shown.
    Desired values are shown in black and measured ones in blue.}
    \label{fig:walking-xB}
    \vspace{-8mm}
\end{figure}
\begin{figure*}[t]
    \centering
    \includegraphics[width = \textwidth]{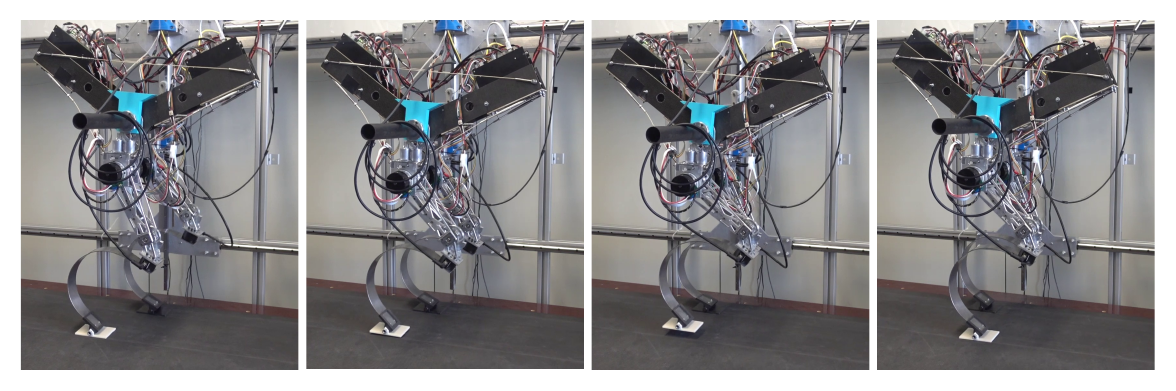}
    \caption{Snapshots of a half-stride of a walking experiment with \ramone{}.
    The sequence starts just before lift-off of the right leg.    }
    \label{fig:snapshots-walking}
    \vspace{-6mm}
\end{figure*}

\begin{table}[b]
    \centering
    \begin{tabular}{|l|c|c|c|c|}
        \hline
        Low-level control & $\vP_{\vla}$ & $2.5\vI$ & $\vP_{\nu}$ & $10\vI$ \\
        \hline
        Virtual model control & $p_x$ & $300$ & $d_x$ & $34.6$ \\
        & $p_y$ & $2000$ & $d_y$ & $89.4$ \\
        \hline
        Gait scheduling &
        $\beta$ & $0.6$ & $T_{\mathrm{str}}$ & \unit[1.2]{s} \\
        & $\theta_L$ & $0$ & $\theta_R$ & $0.5$ \\
        & $\lambda_{y}^{min}$ & \unit[5]{N} & & \\
        \hline
        Force distribution  & $w_x$ & $1$ & $\mu$ & $0.5$\\
        & $\lambda^{max}_{y,L}$ & \unit[350]{N} (0)& $\lambda^{max}_{y,R}$ & \unit[350]{N} (0) \\
        \hline
    \end{tabular}
    \caption{Parameter choices for the walking experiment. 
    }
    \label{tab:control-parameters}
\end{table}

With these parameter choices, stable periodic walking is possible.
The resulting forward and vertical motions over \unit[10]{s} of walking are shown in Fig.~\ref{fig:walking-xB}.
The measured forward motion follows the desired horizontal trajectory with a constant offset of around~\unit[10]{mm}.
The measured height oscillates around the desired height with maximal errors around \unit[40]{mm} over the course of a gait cycle.
This is caused by the abrupt changes in the desired ground reaction forces between the double and single support phases and the limited bandwidth of the low-level force controller.
As a result, the main body looses height at the beginning of the single support phase and moves upward when transitioning from single to double support.
Four snapshots of a half-stride of the walking motion are shown in Figure~\ref{fig:snapshots-walking}.
A video recording of this experiment is provided in the supplementary materials.

\section{Conclusion} \label{sec:conclusion}
This paper introduces a novel type of legged robotic system with continuously compliant structures.
These function as series elastic elements, replacing traditional series elastic actuators (SEA).
The concept is demonstrated using a planar bipedal robot with lower leg segments made from C-shaped spring steel.
We present a suitable model for this design and, building on principles from traditional SEA robots, employ virtual model control to achieve stable walking.
The robot successfully walks at a velocity of \unitfrac[0.15]{m}{s}.

The performance of the proposed system is on par with that achieved using traditional SEAs on the same platform.
Walking experiments reported in~\cite{smit2017ramone} achieved a velocity of \unitfrac[0.2]{m}{s}.
Going beyond the notable simplicity of the proposed controller and incorporating a dynamical model of the robot, a nonlinear model of the spring deflection, and a carefully calibrated force estimation algorithm could potentially enhance the performance of the robotic system further.

While the walking performance was generally robust, we observed deviations from the desired main-body trajectories in our experiments, particularly in the vertical direction.
To address these deviations, it is crucial to incorporate the elastic elements into the trajectory planning for both the main body and legs, as the inherent bandwidth limitations of the high-compliance drivetrain prevent increases in the gains of the virtual model controller.
For example, the abrupt transitions between single and double stance phases in our current gait scheduler lead to jumps in the desired ground reaction forces which cannot be realized on the hardware.
To mitigate this, the gait scheduler could be adjusted to gradually transition weight between single and double stance phases by incrementally reducing the force limits ($\lambda_{y,i}^{max}$) on the swing foot before liftoff, and gradually increasing them upon touchdown.

A primary limitation of this initial work is the restriction to purely sagittal motion imposed by the motion planarizer, which also fixed the pitch angle of the main body. 
This approach was chosen to simplify postural control and to focus on evaluating the effects of the compliant structure. 
Removing the pitch constraint would result in an under-actuated single stance phase, where the main body pivots passively around the ground contact point.
Although we believe that the presented controller could be extended to accommodate this effect, it would require carefully designed motion trajectories.

The primary motivation for this work was to achieve a simplified and more lightweight leg design. 
However, we are not yet able to provide quantitative conclusions regarding reductions in cost, weight, and complexity.
This is because the platform used in our experiments was based on a traditional SEA robot with its elastic elements disabled, meaning we inherited the complexity and weight of the SEA without utilizing its advantages. 
To make a meaningful comparison, we need to develop a dedicated robot from scratch. 
This new robot should explore additional weight-saving strategies, such as using advanced materials like carbon-fiber reinforced plastic for the compliant shanks. 
Additionally, while using off-the-shelf force/torque sensors facilitated rapid evaluation and reduced development time, these sensors added extra mass to the robotic leg. 
A significant weight reduction could be achieved by measuring spring deflection rather than spring forces by, for example, integrating strain gauges directly onto the springs~\cite{bendfeld2024deformation}.

Finally, we have yet to fully explore the potential for energy storage and retrieval during locomotion with our high-compliance actuation system. 
As predicted by various template models~\cite{blickhan1989spring, gan2018all}, leveraging this capability could significantly enhance the versatility and energy efficiency of robotic locomotion.
Future work will focus on optimizing this aspect to maximize the benefits of the compliant leg design.

\bibliographystyle{ieeetr}
\bibliography{References}

\end{document}